\documentclass[10pt,twocolumn,letterpaper]{article}

\usepackage[pagenumbers]{cvpr} % To force page numbers, e.g. for an arXiv version

\usepackage{amsmath,amssymb,amsthm,mathtools}
\usepackage{booktabs,multirow,graphicx,xcolor}
\usepackage{array}
\usepackage{xspace}
\usepackage{enumitem}
\usepackage{subcaption}
\usepackage{url}
\newcommand{\status}[2]{\par\smallskip\noindent\colorbox{%
\ifnum\pdfstrcmp{#1}{FINAL}=0 green!12\else\ifnum\pdfstrcmp{#1}{FINAL (theory)}=0 green!12\else orange!18\fi\fi}{\parbox{\dimexpr\linewidth-2\fboxsep}{\scriptsize\textbf{[#1]}~#2}}\par\smallskip}

\graphicspath{{figures/}{./figures/}{CVPR27_draft_v2_authorkit/figures/}}

\usepackage{tikz}
\usetikzlibrary{arrows.meta,calc}
\usepackage{booktabs,tabularx}

\newtheorem{proposition}{Proposition}

\newcommand{\E}{\mathbb{E}}

\newcommand{\calD}{\mathcal{D}}

\newcommand{\calA}{\mathcal{A}}

\newcommand{\calB}{\mathcal{B}}

\newcommand{\Ghat}{\hat{G}}

\newcommand{\alphahat}{\hat{\alpha}}

\newcommand{\adigen}{\text{ADIGen}\xspace}

\newif\ifshowstatus\showstatustrue
\let\origstatus\status
\renewcommand{\status}[2]{\ifshowstatus\origstatus{#1}{#2}\fi}

\newcommand{\CURE}{\textup{\textsc{Cure}}\xspace}
\newcommand{\dxName}{certificate\xspace}      % the rule, in prose
\newcommand{\DxName}{Certificate\xspace}      % the rule, in prose
\newcommand{\dxScore}{\ensuremath{\mathsf{Cert}}}     % the score, in math: \dxScore(\Ghat,\pi_\beta), \dxScore_b
\newcommand{\kickc}{{kick2c}\xspace}

\newcommand{\LSlong}{low-support\xspace}
\newcommand{\support}{support\xspace}

\newcommand{\Gstd}{\hat G_{\mathrm{standard}}}     % standard loss
\newcommand{\Gdeb}{\hat G_{\mathrm{debias}}}     % debiased loss (ADIGen)
\newcommand{\Grt}{\hat G_{\mathrm{retarget}}}       % debiased + retargeted
\newcommand{\Gset}{\mathcal{G}}               % the candidate set

\newcommand{\Bcon}{\calB_{\mathrm{con}}}
\newcommand{\Bls}{\calB_{\mathrm{ls}}}

\newif\ifshowpending\showpendingtrue

\definecolor{cvprblue}{rgb}{0.21,0.49,0.74}
\usepackage[pagebackref,breaklinks,colorlinks,allcolors=cvprblue]{hyperref}

\def\confName{CVPR}
\def\confYear{2027}

\newcommand{\titleString}{Plausible Videos Can Lead to Poor Decisions:\\Diagnosing and CURE-ing World Models for Planning}
\title{\titleString}

\author{
Raphael C. Kim$^{1}$ \quad
Michele Santacatterina$^{2}$
Jingsen Zhu$^{1}$ \quad
Ramin Zabih$^{1}$ \\
$^{1}$Cornell University \quad
$^{2}$NYU Grossman School of Medicine\\
{\tt\small rk853@cornell.edu}
}

\begin{document}

\maketitle

\begin{abstract}
Action-conditioned world models are increasingly used as world models for planning, yet a world model that makes plausible, realistic-looking videos that are consistent with the training data can produce poor decisions.
We address two key questions:
(1) when can we trust a world model for planning, and (2) 
what should we do when we cannot? 
We introduce \CURE (\textbf{C}ertify, \textbf{U}pweight, \textbf{R}ollout, \textbf{E}xplore), a certificate-guided technique that assesses whether a world model can be trusted for planning and determines what to do next: whether to upweight the training data to debias the model; roll out more samples at test time; or explore by collecting new data where \support (how well the training data cover the planner's decisions) is missing.
We deploy \CURE across multiple domains, spanning simulated physics based on PhyWorld, MuJoCo arm control, and Language-Table pushing in simulation and on real videos.
The decisions our certificate trusts have lower regret than those it does not (e.g., $0.06$ vs.\ $0.38$ on Language-Table sim and $0.07$ vs.\ $0.38$ on MuJoCo contact), whereas video perceptual quality does not identify good decisions.
By focusing on decision quality rather than perceptual quality, \CURE lowers regret over the best pessimistic decision rule by 10--25\% across \support levels, and over world-action models, search and policy learning by 41--64\%, given the same data.
\end{abstract}

\section{Introduction}\label{sec:intro}

Video generators are widely used as world models: condition on what the camera has seen and an action of interest, then roll out the generator to see which action to take~\cite{yang2023learning,brooks2024video,bruce2024genie,cogvideox2024,cosmos2025}.
The reasoning seems defensible: a correct world model should produce convincing video, so convincing video may provide evidence that a world model predicts the consequence of an action well.
However, there is considerable evidence that this is not the case: video generators that produce such videos can still violate basic physics~\cite{phyworld2024,physicsiq2025,videophy2024,videophy2_2025,physion2021} and they do not necessarily lead to good decisions~\cite{worldscore2025,worldsimbench2024,worldmodelbench2025,yang2024ewm}.

\begin{figure}[!t]
  \centering
  \includegraphics[width=\linewidth]{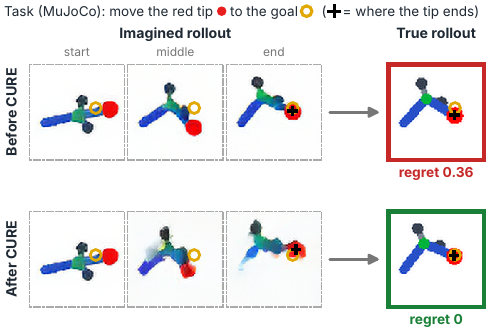}
  \caption{\textbf{Convincing video does not mean a better action.}
  World models make decisions by generating videos as imagined rollouts, predicting the regret from the videos, and picking the action with the lowest predicted regret. A more convincing video may still predict regret badly, and choose an action that fails in the true rollout (top). A less convincing video, despite the visible artifacts (ghosting, blurring and smearing of the arm), can predict accurately and select a better action (bottom).}
  \label{fig:teaser}
\end{figure}

Fig.~\ref{fig:teaser} shows a concrete example of the potential disagreement between video quality and decision quality. Here, the planner needs to move the red tip on a robot arm toward the target position (the MuJoCo benchmark~\cite{todorov2012mujoco}, details in \cref{sec:setup_exp}), and a world model is trained to generate a video of the robot arm's movement under a specific action, i.e., an \emph{imagined rollout}. 
For each candidate action, the model predicts its regret by comparing the red tip's final position in the imagined rollout with the target position, and the planner favors the action with lower predicted regret. However, in our example, the first row shows a model generating a more convincing video whose imagined rollout leads to the wrong decision. The second row shows the same scene after applying \CURE, the technique we introduce in this paper. The imagined rollout is visibly less convincing, with a blurred and ghosted arm, yet the action the planner selects from it is the best one available (regret $0$).
Nothing in the first row signals the problem: the video is sharp and shows the tip reaching the goal.
This is not an isolated case: across decisions and domains, video quality does not identify the decisions with low regret (\cref{sec:trust_results}).

Why does this happen? 
Prior work suggests several reasons. At test time, a planner may score actions on too few rollouts being misled by noise~\citep{kearns2002sparse,kim2026cosmospolicy}, or search too few candidate actions, missing better ones~\citep{ebert2017bair,hafner2019planet,ma2025inference,liu2025videot1}. 
The model may also lack the capacity to learn the dynamics~\citep{hansen2024tdmpc2}. 
Finally, the training data may teach correlations rather than general laws, so the model fails to perform out of distribution~\citep{dehaan2019causal,phyworld2024,kim2026adigen}.
These explanations for planning failure reduce to two quantities: model error and \support (how well the training data covers the planner's decisions). Indeed, planning error, or regret, is bounded by the product of these two quantities, plus regularization bias~\cite{jin2021pessimism, uehara2022pessimistic, kim2026adigen}.
Standard world model training fits the training data and rewards visually compelling videos, but controls neither error source. As a result, a world model can produce convincing video, but plan poorly. 

Existing solutions focus on different error sources.
Scaling test-time compute, with more rollouts~\cite{kearns2002sparse,kim2026cosmospolicy} or more candidates~\cite{ebert2017bair,hafner2019planet,ma2025inference}, reduces rollout noise and error. However, if \support is low, the model's error cannot be fixed unless \support is addressed. 
Debiased training and retargeting~\cite{kim2026adigen} lowers the model's error where the training data has a few clips of the rarely recorded pairs, but cannot help when there is effectively no \support.
Collecting more data helps, but only if the new clips are collected where \support is missing: more data from the same training source is limited by coverage, not its size~\cite{levine2020offline,yu2020mopo,kidambi2020morel}.
Exploration that targets the model's uncertainty~\cite{houlsby2011bald,pathak2017curiosity,shyamsundar2019active} could find those gaps, but it relies on uncertainty estimates that are themselves unreliable out of distribution~\cite{ovadia2019trust,nalisnick2019deep}.

Each solution addresses a different root cause, and sometimes none is needed. 
We address two fundamental questions:
\begin{enumerate}[label=\textbf{(Q\arabic*)},leftmargin=2.2em,itemsep=1pt,topsep=2pt]
  \item \emph{When can we trust our world model for planning?}  \label{q:when}
  \item \emph{When we cannot trust our model for planning, what can we do about it?} %Which solution does it need: a debiased loss, more compute, or more data, and where? 
  \label{q:what}
\end{enumerate}

\vspace{-1em}

\paragraph{Contributions.}
We introduce \CURE (\textbf{C}ertify, \textbf{U}pweight, \textbf{R}ollout, \textbf{E}xplore).
\begin{enumerate}[itemsep=1pt,topsep=2pt]
\item \textbf{When to trust a world model: Certify.}
We produce a theoretically justified certificate for when to trust a world model for planning. The certificate is easily computed from the training data, using estimates of the model's error and of its \support. 
It guarantees the planner is as good as any fixed action, up to these two quantities, connecting each error source to planning error.
\item \textbf{What to do about it: Upweight, Rollout, Explore.} We introduce \CURE, a unified technique that turns the certificate into go/no-go criteria for fixing a world model until its decisions pass the certificate (Fig.~\ref{fig:tree}). These criteria say when to upweight the training data to debias the model, when to roll out more samples, and when, and where, to explore by collecting new data.
% \item \textbf{Experimental evidence across multiple domains.} On simulated physics, robot arm control, and tabletop pushing in simulation and on real videos, the certificate separates good decisions from bad whereas video quality does not (\cref{fig:trust}), and given the same data and compute, \CURE lowers regret over the strongest competing method by 41--61\% across \support levels (\cref{tab:strategy}). Overall, \CURE lowers regret over standard world models by 31--61\% where \support is low and by up to 50\% where it is partial, while never doing worse where \support is high.
\item \textbf{Experimental evidence across multiple domains.} On simulated physics, robot arm control, and tabletop pushing in simulation and on real videos, the certificate separates good decisions from bad whereas video quality does not (\cref{fig:trust}). 
Given the same data, \CURE lowers regret over the best pessimistic decision rule by 10--25\% across \support levels, and over world-action models, search and policy learning by 41--64\% (\cref{tab:strategy}).
\end{enumerate}

\section{Related Work}\label{sec:related}

\paragraph{Video world models and their evaluation.} Video models are increasingly used for planning by imagining each candidate action's outcome~\cite{finn2017deep,ebert2018visual,du2023unipi,yang2023learning,bruce2024genie,genie3_2025,cosmos2025}. Benchmarks grade the generated video itself~\cite{vbench2024,videophy2_2025,physicsiq2025,phyworld2024,worldmodelbench2025}, yet visual quality does not guarantee task success~\cite{zhang2026worldinworld}. Policy evaluators~\cite{quevedo2026worldgym,gemini2025veo} and action-following benchmarks~\cite{chen2026worldecho} judge a model with something outside its training data, such as simulators or real trials. None asks whether, \emph{from the training data alone}, a model can be trusted to rank actions, or what to do when it cannot.

\paragraph{Policy learning with world models.} Policies are learned by planning in a world model~\cite{hansen2024tdmpc2} or training on its rollouts~\cite{sutton1991dyna,janner2019trust,hafner2025dreamerv3,alonso2024diamond,zhu2025wmpo}. Online methods collect fresh data~\cite{feng2023fowm,guo2026vlaw} or simulated outcomes of non-expert actions~\cite{yang2025resim,pandya2026picowam} where the policy exploits model errors. Offline methods are limited by how well the training data covers the best policy~\cite{rashidinejad2021bridging,uehara2022pessimistic}, and stay close to the data~\cite{kostrikov2022iql,kumar2020cql,yu2021combo} or penalize model uncertainty~\cite{yu2020mopo,kidambi2020morel}. Uncertainty penalties trust the model to know where it is wrong, but generative models can be confident far from their data~\cite{nalisnick2019deep,lu2022revisiting}. We instead treat \support as a property of the training data, measured with importance weights.

\paragraph{Debiased estimation and safe policy improvement.} Doubly robust estimation corrects for the behavior policy by combining an outcome model with importance weights~\cite{robins1994estimation,dudik2011doubly,chernozhukov2018double,chernozhukov2022automatic}, and recent work moves such corrections into model training~\cite{hishinuma2021weighted,luedtke2025doublegen,melnychuk2026gdr}. We adopt \adigen~\cite{kim2026adigen} as our debiasing step. Where weights grow large, estimators cap them and fall back on the model~\cite{wang2017switch,su2019cab}, and where the behavior policy never takes an action, there is nothing to reweight~\cite{sachdeva2020deficient}: debiasing repairs \LSlong but not absent \support. Safe policy improvement accepts a new policy only as far as the training data supports it~\cite{swaminathan2015crm,thomas2015hcpi,laroche2019spibb}. \CURE brings this test to world models, and collects data where it fails.

\paragraph{Imagination and data collection.} Retraining a generator on its own rollouts adds no evidence and risks model collapse~\cite{shumailov2024ai,alemohammad2023self}, so missing \support must be filled with executed clips. Active learning and exploration collect where the model is uncertain~\cite{houlsby2011bald,pathak2017curiosity,shyamsundar2019active,sekar2020plan2explore}, which misleads when a generator is confident without data. Our collection rule targets missing \support instead.
\section{Setup and Background}\label{sec:setup}

\subsection{Problem}\label{sec:problem}

The objective is to learn a world model from training data, and use the world model for planning.
Our data are $n$ i.i.d. clips $(X_i,A_i,Y_i)$, comprised of a context $X$ or the conditioning frames from a video, an action $A\in\calA$, and an outcome video $Y$ with utility $U(Y)$, under a behavior policy $\pi_b(a\mid x)$.
A world model $G$ is a conditional video model: it takes a scene and an action as input to produce a generated video, approximating $P(Y\mid X=x,A=a)$. We assume standard identifiability conditions~\citep{pearl2009causality} so that this equals the interventional law $P(Y(a)\mid X=x)$, where $Y(a)$ is the video that would result from setting $A=a$. 
Planning chooses one action, rolled out over the whole video.

A planner $\pi$ is a map from scenes to action distributions over an action set $\calB\subset\calA$. With $\mu(x,a)=\E[U(Y(a))\mid X=x]$ the expected utility of action $a$ and $a^\star(x)=\arg\max_{a\in\calB}\mu(x,a)$ the best action, we minimize regret:
\begin{align*}
  \mathrm{Reg}(\pi)=\E_X\Big[\mu\big(X,a^\star(X)\big)-\sum_{a\in\calB}\pi(a\mid X)\,\mu(X,a)\Big].
\end{align*}
A summary of the notation used throughout the paper is provided in the Supplement.

\subsection{Planners}\label{sec:planners}

We plan with an estimated utility function $\hat u(x,a)=\E\big[U(\Ghat(x,a))\big]$ derived from rollouts under the estimated world model $\Ghat$, and with $\hat\pi_b$, a probability distribution over $\calB$ that estimates $\pi_b$.
The primary planner we use is KL-regularized, so that decisions poorly supported by our data are penalized $\pi_\beta(a\mid x)\propto\hat\pi_b(a\mid x)\exp\big(\hat u(x,a)/\beta\big) $. 
$\beta\to0$ yields the greedy choice $\arg\max_{a\in\calB}\hat u(x,a)$, while $\beta\to\infty$ gives the fitted behavior policy $\hat\pi_b$.
KL regularization is widely used~\cite{peters2010reps,peng2019awr,nair2020awac,wu2019brac,levine2020offline,janner2022diffuser,ziegler2019finetuning}. We additionally evaluate greedy, advantage-weighted~\cite{peng2019awr}, behavior-cloning-regularized~\cite{fujimoto2021td3bc}, batch-constrained~\cite{fujimoto2019bcq} and baseline-bootstrapped~\cite{laroche2019spibb} planners. We also compare against the behavior policy's default action, search with a verifier, the rules that world-action models use~\cite{du2023unipi,pandya2026picowam}, and pessimistic rules that score each action by a lower bound on its utility (the standard remedy for limited \support~\cite{jin2021pessimism,yu2020mopo,kidambi2020morel,lakshminarayanan2017simple}).

\subsection{Three Solutions for Building Decision-Ready World Models}\label{sec:solutions}

When a world model plans poorly, three solutions are available --- each targets a different root cause of planning error:

\paragraph{Upweight: Debiased training and retargeting.}\label{sec:adigen}
A \textbf{standard} world model fits the training data as is, so it learns the scene--action pairs it sees often well and extrapolates poorly under \LSlong. The \textbf{debiased} loss of \adigen~\cite{kim2026adigen} corrects this with a weight $\alphahat(x,a)$ estimating the density ratio $\nu(a\mid x)/\pi_b(a\mid x)$ between a uniform interventional policy $\nu$ over $\calB$ and the behavior policy $\pi_b$.
Rather than divide by an estimate of $\pi_b$, which is unstable, the weight is `automatically' learned by directly fitting a Riesz regression~\cite{chernozhukov2022automatic} on the training data $\calD_0$, and is capped at $M$ for stability --- the full loss is restated in the supplementary material.
\textbf{Retargeting}~\citep{kim2026adigen} then continues training with $\nu$ replaced by the planner's own action distribution, so that the model is most accurate where the planner acts.

\paragraph{Rollout: Scaling test-time compute.} At decision time, test-time compute can be scaled in two ways: scoring more candidate actions, or sampling more rollouts per candidate.
For the first, draw $K$ candidates from $\calB$, roll out each, and choose the one with the highest $\hat u$~\cite{ebert2017bair,hansen2024tdmpc2,ma2025inference}.
This can improve planning, but a larger $K$ makes model over-optimism more likely, especially on actions the training data rarely contains.
For the second, more rollouts per candidate reduce the variance of $\hat u(x, b)$ but not its bias.
Both help only when the model is accurate on the actions being compared.

\paragraph{Explore: Data collection.} The alternative is to execute an action, add the clip to the data, and retrain the world model.
Given a budget of $m$ clips, a collection rule maps the current model, training data, and pool of scenes to $m$ scene--action pairs to execute, giving a dataset to augment with, $\calD_m$.

\section{\CURE: When to Trust a World Model, and What to Do About It}\label{sec:method}

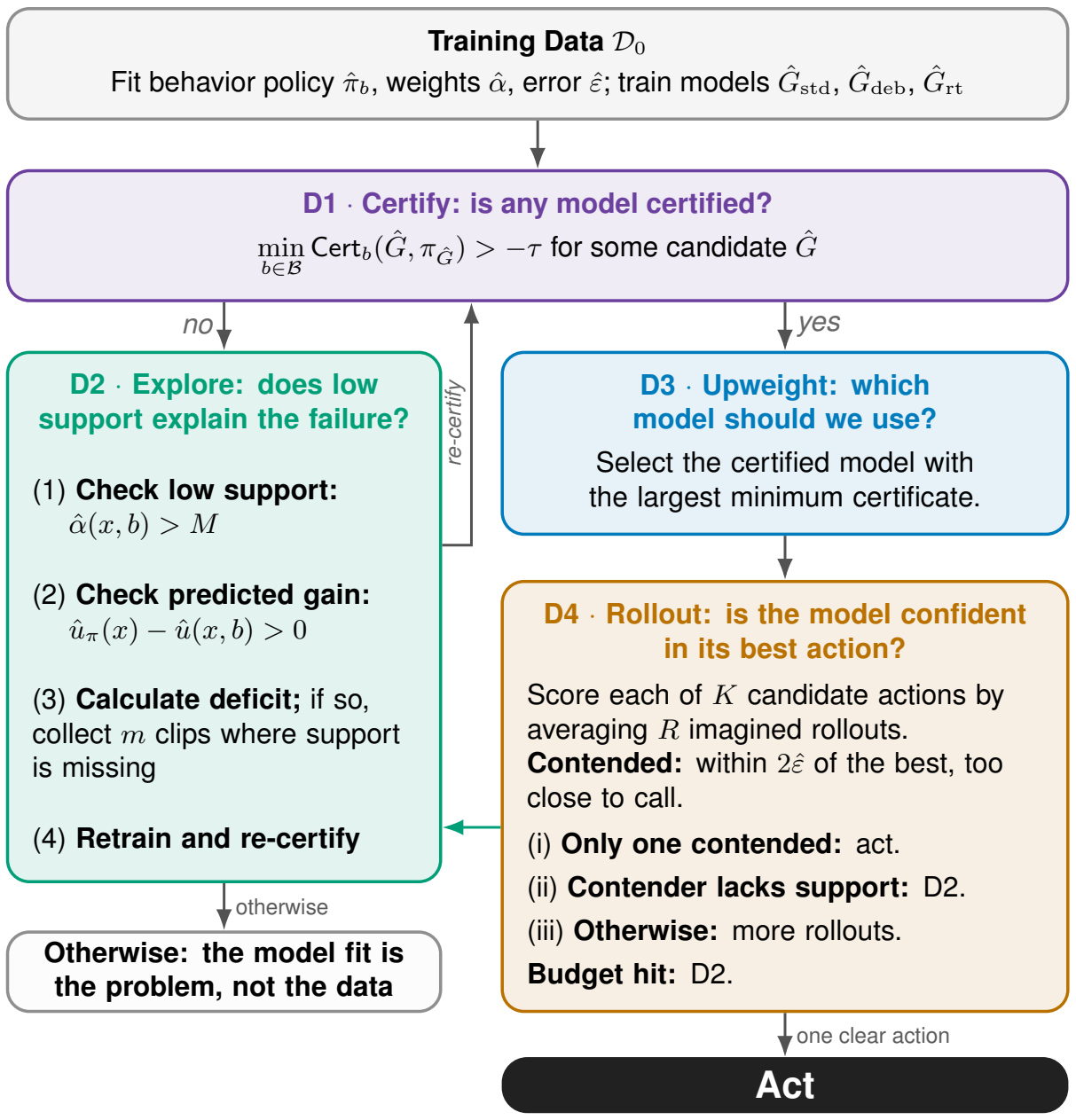
\begin{figure}[t]
  \centering
  \resizebox{\linewidth}{!}{%
    % CURE decision tree only: include inside the manuscript's figure environment.
% Required in preamble: \usepackage{tikz,graphicx,xcolor}
% \usetikzlibrary{arrows.meta,calc}
% Use: \resizebox{\linewidth}{!}{\input{figures/fig_tree_tikz}}
\definecolor{cureFigGray}{HTML}{8F8F8F}
\definecolor{cureFigPurple}{HTML}{6B409E}
\definecolor{cureFigGreen}{HTML}{009F78}
\definecolor{cureFigBlue}{HTML}{007BBB}
\definecolor{cureFigOrange}{HTML}{B97100}

\begin{tikzpicture}[
  >=Latex,
  font=\sffamily\small,
  frame/.style={rounded corners=7pt, line width=1pt, align=center,
                inner xsep=8pt, inner ysep=7pt},
  colL/.style={frame, text width=4.34cm, minimum width=4.9cm, anchor=north},
  colR/.style={frame, text width=5.84cm, minimum width=6.4cm, anchor=north},
  arrow/.style={-Latex, line width=.85pt, draw=black!68},
  back/.style={-Latex, line width=.85pt, draw=cureFigGreen},
  lab/.style={font=\sffamily\itshape\small, text=black!60},
]
% columns: left 0.15-5.05 cm, right 5.75-12.15 cm (0.7 cm gutter centred at 5.40)
\def\XL{2.60}\def\XR{8.95}\def\XM{5.40}

% Training data and candidate world models
\node[frame, draw=cureFigGray, fill=black!4, minimum width=12.0cm] (log) at (6.15,0) {
  {\bfseries Training Data $\calD_0$}\\[2pt]
  Fit behavior policy $\hat\pi_b$, weights $\hat\alpha$, error $\hat\varepsilon$;
  train models $\hat G_{\mathrm{std}}$, $\hat G_{\mathrm{deb}}$, $\hat G_{\mathrm{rt}}$
};

% D1
\node[frame, draw=cureFigPurple, fill=cureFigPurple!10, minimum width=12.0cm,
  anchor=north] (d1) at ($(log.south)+(0,-.55)$) {
  {\color{cureFigPurple}\bfseries D1 $\cdot$ Certify: is any model certified?}\\[3pt]
  $\displaystyle\min_{b\in\calB}\dxScore_b(\Ghat,\pi_{\Ghat})>-\tau$
  for some candidate $\Ghat$
};

% Right column: D3 -> D4 -> Act
\node[colR, draw=cureFigBlue, fill=cureFigBlue!10] (d3) at (\XR,{-3.25}) {
  {\color{cureFigBlue}\bfseries D3 $\cdot$ Upweight: which model should we use?}\\[3pt]
  Select the certified model with the largest minimum certificate.
};
\node[colR, draw=cureFigOrange, fill=cureFigOrange!11, align=left]
  (d4) at ($(d3.south)+(0,-.5)$) {
  \centering
  {\color{cureFigOrange}\bfseries D4 $\cdot$ Rollout: is the model confident in its best action?}\\[4pt]
  \raggedright
  Score each of $K$ candidate actions by averaging $R$ imagined rollouts. \textbf{Contended:} within $2\hat\varepsilon$ of the best, too close to call.\\[4pt]
  (i) \textbf{Only one contended:} act.\\[3pt]
  (ii) \textbf{Contender lacks \support:} D2.\\[3pt]
  (iii) \textbf{Otherwise:} more rollouts.\\[3pt]
  \textbf{Budget hit:} D2.
};

% Fallback: bottom aligned with D4
\node[frame, draw=cureFigGray, fill=black!1, text width=4.34cm, minimum width=4.9cm,
  minimum height=.62cm, inner ysep=3pt, font=\sffamily\small\bfseries, anchor=south] (fallback) at (\XL,0 |- d4.south) {
  Otherwise: the model fit is the problem, not the data};

% Left column: D2 fills the space between D1 and the fallback, steps spread evenly
\path let \p1=(d3.north), \p2=($(fallback.north)+(0,.5)$) in
  node[colL, draw=cureFigGreen, fill=cureFigGreen!11, align=left, inner ysep=0pt]
  (d2) at (\XL,\y1) {%
  \begin{minipage}[c][\dimexpr\y1-\y2-2pt\relax][t]{4.34cm}
  \vspace{7pt}\centering
  {\color{cureFigGreen}\bfseries D2 $\cdot$ Explore: does low support explain the failure?}\par
  \raggedright\vfill
  (1) \textbf{Check low support:}\\ \hspace*{1.3em}$\hat\alpha(x,b)>M$\par\vfill
  (2) \textbf{Check predicted gain:}\\ \hspace*{1.3em}$\hat u_\pi(x)-\hat u(x,b)>0$\par\vfill
  (3) \textbf{Calculate deficit;} if so, collect $m$ clips where support is missing\par\vfill
  (4) \textbf{Retrain and re-certify}\par\vspace{9pt}
  \end{minipage}%
};

% Terminal nodes on one shared row
\node[rounded corners=8pt, draw=black!87, fill=black!87, text=white,
  minimum width=6.4cm, minimum height=.62cm, font=\sffamily\bfseries\large,
  anchor=north] (act) at ($(d4.south)+(0,-.5)$) {Act};

% Branches
\draw[arrow] (log.south) -- (d1.north);
\draw[arrow] (d1.south -| d2.north) -- node[left, lab] {no} (d2.north);
\draw[arrow] (d1.south -| d3.north) -- node[right, lab] {yes} (d3.north);
\draw[arrow] (d3.south) -- (d4.north);
\draw[arrow] (d2.south) -- node[right, font=\sffamily\scriptsize, text=black!65] {otherwise} (fallback.north);
\draw[arrow] (d4.south) -- node[right, font=\sffamily\scriptsize, text=black!65] {one clear action} (act.north);

% D2 -> D1: collect real clips, retrain, re-certify (up the centre gutter)
\coordinate (mid) at (\XM,0);
\coordinate (rc) at ($(d2.north east)+(0,-2.2)$);
\draw[arrow] (rc) -- (rc -| mid) -- node[midway, sloped, above, inner sep=1.5pt,
  font=\sffamily\itshape\scriptsize, text=black!60] {re-certify} (mid |- d1.south);

% D4 -> D2: contended low-support action, or rollout budget exhausted
\coordinate (bk) at ($(d4.west)+(0,-.35)$);
\draw[back] (bk) -- (bk -| d2.east);
\end{tikzpicture}
  }
\caption{\textbf{\CURE.} We estimate the behavior policy $\hat\pi_b$ and the weights $\hat\alpha$ on the training data, and the model error $\hat\varepsilon$ on a validation set. We then train the candidate models (standard, debiased, retargeted) and certify each one (D1). If none passes, we collect new clips where the certificate fails and re-certify, until the clip budget is spent (D2). We keep the model with the largest certificate margin (D3), and spend extra rollouts only on contended actions, those too close to the top score to tell apart, that have \support (D4). Full procedure in the supplementary material.}
  \label{fig:tree}
\end{figure}

Each solution of \cref{sec:solutions} targets a different root cause, and sometimes none is needed. 
We first derive a \dxName for assessing whether a world model can be trusted for planning, answering \labelcref{q:when}, and then introduce \CURE, a certificate-guided framework that determines what to do next, answering \labelcref{q:what}.

%--------------------------------------------------------------
\subsection{The \DxName}\label{sec:certificate}

Throughout, for any $f$ that acts on clips and actions, we let $\|f\|^2=\E_{\calD_0}[f(x,a)^2]$. We require two pieces: a weight measuring \support, and an estimate of model error.
First, for a planner $\pi$ over $\calB$, the weight $\alpha_\pi(x,a)=\pi(a\mid x)/\pi_b(a\mid x)$ is the density ratio between $\pi$ and the behavior policy, measuring how far $\pi$ is from the behavior policy.
Second, the model's error for planning is $\varepsilon=\|\bar\delta\|$, where $\delta(x,a)=\mu(x,a)-\hat u(x,a)$ and $\bar\delta$ centers $\delta$ over the actions in $\calB$.
Together, they bound the regret of a planner built from a world model:
\begin{proposition}[Restated from \citet{kim2026adigen}]\label{prop:dx}
For any planner $\pi$ over $\calB$, let $\pi^\star$ be the optimal planner that always takes $a^\star(x)$, and $\hat u_\pi(x)=\sum_{b\in\calB}\pi(b\mid x)\,\hat u(x,b)$ the model's predicted utility of $\pi$. Then,
\begin{equation*}
  \mathrm{Reg}(\pi)\leq\varepsilon\,\|\alpha_{\pi^\star}-\alpha_{\pi}\|+\E\big[\hat u(x,a^\star)-\hat u_\pi(x)\big].
\end{equation*}
\end{proposition}
The first term combines the model's centered utility error on the training distribution, $\varepsilon$, with the difference between the optimal and learned planners relative to the behavior policy.
This term can be large when the model is inaccurate or when the planners differ on actions poorly supported by the training data.
The second term measures how much predicted utility we lose by following the learned planner instead of the optimal action. For our KL-regularized planner, this can reflect regularization bias.

\paragraph{From the regret bound to a \dxName.}
A small bound in \cref{prop:dx} would support trusting the planner, but the bound depends on the optimal policy $\pi^\star$, which is unknown.
We therefore apply the same reasoning to each fixed action $b\in\calB$: how much do we lose by following our planner instead of always taking $b$? 
Equivalently, we lower-bound the improvement: \textit{how much might our planner improve over a constant-action planner $b$?}

To answer this, we substitute estimates into \cref{prop:dx}: the weights $\hat\alpha$ follow from the fitted behavior policy $\hat\pi_b$, and $\hat\varepsilon$ is estimated on validation clips: a cross-fitted regression \citep{chernozhukov2018double} of the residual $U(Y)-\hat u(X,A)$ on scene and action gives a bias $\hat\delta(x,b)$, which we center over $\calB$. 
This gives the certificate for action $b$:
\begin{equation}\label{eq:certificate}
  \dxScore_b(\Ghat,\pi)=\E\big[\hat u_\pi(x)-\hat u(x,b)\big]-\hat\varepsilon\,\|\hat\alpha_b-\hat\alpha_\pi\|,
\end{equation}
The first term is how much the model predicts our planner will improve over $b$. 
The second accounts for how much model error and limited support could change this comparison.  
If $\dxScore_b>0$, the estimated bound favors our planner over $b$, even after accounting for model error and limited support. If this holds for every $b\in\calB$ up to a tolerance $\tau$, that is, $\min_{b\in\calB}\dxScore_b(\Ghat,\pi)>-\tau$, we \emph{certify} our world model: no fixed action is estimated to outperform our planner by more than $\tau$. This certification holds with high probability (Supplement).

\paragraph{Per decision.} For a single scene $x$, dropping the expectation gives $\dxScore_b(x)=\hat u_\pi(x)-\hat u(x,b)-\hat\varepsilon\,\|\hat\alpha_b-\hat\alpha_\pi\|_x$
with the norm over actions at $x$. If $\dxScore_b(x)>-\tau$ for every covered $b$ ($\hat\alpha_b\leq M$), the decision has regret at most $\tau$ (Supplement); \cref{sec:trust_results} uses this.

When the certificate fails, the pieces indicate why: the model's error is too large, the planner strays too far from the \support of the training data, or the predicted gain is too small. Each root cause calls for a different fix, which \CURE makes precise. 

%% Theory finite-sample, appease C.

\subsection{From Certificate to Actions}\label{sec:guidelines}
We summarize our pipeline in \cref{fig:tree}, read from top to bottom.
We use three candidate world models $\Gset=\{\Gstd,\Gdeb,\Grt\}$ of \cref{sec:solutions} (standard, debiased, and debiased + retargeted) for planning.
We first check whether any world model can be certified (D1). If none is certified, we check whether low support calls for collecting new clips (D2). Otherwise, we select the certified model with the largest margin (D3). Finally, for each scene, we decide whether to act, use more rollouts, or collect new clips (D4).

\paragraph{D1 (Certify): Certifying a world model.} Score every candidate world model. A world model is \emph{certified} when $\dxScore_b>-\tau$ for every $b\in\calB$: no fixed action can beat the planner by more than $\tau$, even after accounting for the model error and \support. If no candidate is certified, the training data cannot support the decision.\\
\emph{Rule:} if $\min_{b\in\calB}\dxScore_b(\Ghat)\leq-\tau$ for every $\Ghat\in\Gset$ and clip budget remains, go to D2; otherwise go to D3.

\paragraph{D2 (Explore): Collect $m$ clips, $\calD_m$, where the certificate fails.}
Low support may explain the failure in D1: when the planners differ on poorly supported actions, $\|\hat\alpha_b-\hat\alpha_\pi\|$ can be large, so the predicted improvement may not be enough to account for model error and limited support.
In D2, we propose to: \textit{(1) Check low support:}
Identify actions $b$ with $\alphahat(x,b)>M$; we denote this set of actions by $\Bls(x)$ (defined in the Supplement).
\textit{(2) Check predicted utility:} Check whether $\hat u_\pi(x)-\hat u(x,b)>0$ for these actions, i.e., the model predicts a gain over $b$ that the training data cannot certify.
\textit{(3) Calculate the deficit:} An action leaves $\Bls(x)$ once its weight falls back under $M$, which for the uniform target $\nu$ means $\hat\pi_b(b\mid x)\geq 1/(|\calB|M)$. The \emph{deficit} counts how many of the $n$ training clips are missing to reach that level,
\begin{align*}
  d(x,b)=n\,\Big[\tfrac{1}{|\calB|M}-\hat\pi_b(b\mid x)\Big]_+ ,
\end{align*}
so it is zero for supported actions and largest, about $n/(|\calB|M)$ clips, for actions the behavior policy never took.
\textit{(4) Collect and retrain:}
If the check in (2) passes, draw $m$ pairs $(x,b)$, $b\in\Bls(x)$, in proportion to $d(x,b)$, execute each, and retrain with $\calD_m$.\\
\emph{Rule:} If some $b\in\Bls(x)$ has $\hat u_\pi(x)-\hat u(x,b)>0$, the model predicts a gain over a \LSlong action that the training data cannot certify: collect $\calD_m$ there, and return to D1.

\paragraph{D3 (Upweight):} Among the certified $\Gstd$, $\Gdeb$, and $\Grt$, which should we choose? They all share the same architecture, training data, training procedure.\\
\emph{Rule:} Among the certified world models $\mathcal{C}\subseteq\Gset$, keep the one with the largest certified margin, that is,
\begin{align*}
  \Ghat^\star=\arg\max_{\Ghat\in\mathcal{C}}\ \min_{b\in\calB}\dxScore_b\big(\Ghat,\pi_{\Ghat}\big)
\end{align*}
This rule says whether the training data's \support suffices ($\Gstd$), whether \LSlong can be corrected without more data ($\Gdeb$), and whether training should follow the planner to where it acts ($\Grt$).

\paragraph{D4 (Rollout): Test-time compute?}
With $\Ghat^\star$ selected in D3, we now ask whether the model can clearly distinguish between actions for a given scene $x$.
For a scene $x$ with $K$ candidate actions, each scored by averaging $R$ rollouts, and preferred action $\hat a(x)=\arg\max_b\hat u(x,b)$, the \emph{contended} actions $\Bcon(x)$ are those the model cannot separate from $\hat a(x)$ within its model error, $\hat u(x,\hat a(x))-\hat u(x,b)\leq 2\hat\varepsilon$.
In D4 we therefore propose to: \textit{(1) Check contended actions:}
If $|\Bcon(x)|=1$, the preferred action $\hat a(x)$ can be clearly distinguished from other actions within the model error, and therefore, we can select the preferred action.
\textit{(2) Check low support:} If $\Bcon(x)\cap\Bls(x)\neq\emptyset$, the contended actions include actions poorly supported by the training data, so we need to return to D2 to collect more data in support for them.
\textit{(3) Rollout:} Otherwise, double the rollouts on the contended actions ($R\to 2R$) and the number of candidate actions ($K\to 2K$, adding only supported actions), recompute $\Bcon(x)$, and repeat until (1) or (2) applies or the compute budget is spent; if still undecided, return to D2.\\
\emph{Rule:} Double the rollouts per contended action ($R \to 2R$) and the number of candidate actions ($K \to 2K$, adding only actions with \support), then recompute $\Bcon(x)$. Repeat until one action remains or the compute budget is spent. If still undecided, send $x$ to D2.

\begin{figure*}[t]
  \centering
  \includegraphics[width=\textwidth]{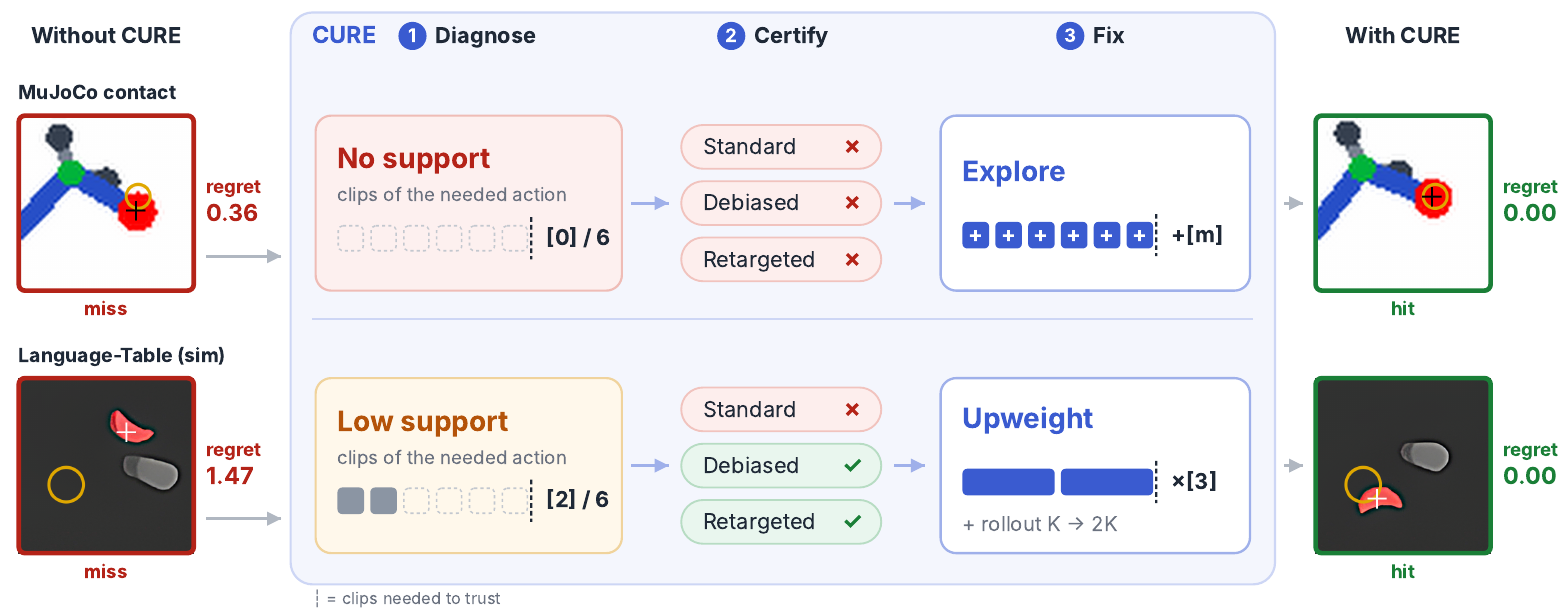}
  \caption{\textbf{\CURE in action.}
  Top: MuJoCo contact. Bottom: Language-Table sim. Left: the standard world model's chosen action misses the goal in the real rollout (true outcome shown).
  Middle: \CURE's certificate decides the fix. Top: the certificate is below zero because the training data has too little \support for the needed action, so \CURE collects new clips (Explore). Bottom: the certificate passes, so \CURE acts, with the upweighted model and more rollouts per action (Upweight, Rollout).
  Right: \CURE's chosen action reaches the goal. The regret numbers are normalized such that 0 is the best action in the bank, and 1 is a random action on average.}
  \label{fig:cure_action}
\end{figure*}

\paragraph{Examples.} \Cref{fig:cure_action} walks through \CURE on two test scenes.
\section{Experiments}\label{sec:experiments}

We test each step of \CURE: Certify (\cref{sec:trust_results}), Explore (\cref{sec:collect_results}), Upweight (\cref{sec:upweight}) and Rollout (\cref{sec:compute_results}). In \cref{sec:strategy_results} we run experiments where we follow all four calls of \CURE, and compare \CURE with other families.

\subsection{Setup}\label{sec:setup_exp}

\paragraph{Domains.} We test \CURE in settings commonly used to evaluate physical prediction and control.
\kickc~\cite{kim2026adigen} is a simulated physics task closely related to PHYRE~\cite{bakhtin2019phyre} and PhyWorld~\cite{phyworld2024}, varied to control how far decisions fall out of distribution: a 2D ball is kicked toward a goal line while a gust, varying from scene to scene, pushes it sideways.
\emph{MuJoCo}~\cite{todorov2012mujoco}, a standard benchmark for model-based control, is a planar arm in two variants: (i) in \emph{contact}, the arm chooses a push to move a block onto a goal spot, a contact-rich setting where outcomes change abruptly with the action; (ii) in \emph{smooth}, the same arm reaches for a goal point with no block, serving as a control where outcomes vary smoothly. 
\emph{Language-Table}~\cite{lynch2023interactive}, widely used to train and evaluate video world models~\cite{yang2023learning}, is a tabletop task in which a 2D push moves a named block toward a target, in simulation (\emph{sim}) and on real robot video (\emph{real}). We simplify by focusing on the continuous action without language. On real video untaken actions cannot be executed, so their outcomes are scored by a separate model. 
Figures in this section focus on Language-Table sim; \cref{tab:strategy} summarizes results over different domains with further details in the supplement.

\paragraph{Support.} To simulate training data with \LSlong, we control how often the behavior policy takes the actions the test scenes need. In each task the behavior policy has a default action for the common case, such as the usual kick under the common gust. A fraction $\eta$ of its clips instead takes other actions inducing \LSlong. We control \support to separate the failure modes it causes, and test the findings in both simulated and real-robot data.
We call $\eta=0.25$ high \support, $\eta=0.05$ partial \support and $\eta=0.005$ low \support. 

\paragraph{Models.} The primary world model is a 28-million-parameter latent video Diffusion Transformer (DiT)~\citep{peebles2023scalable}, the architecture behind today's leading video generators and world models~\citep{brooks2024video,cogvideox2024,cosmos2025}. We call this DiT, trained on the training data with the unweighted loss, the \emph{standard} model; every comparator that needs a world model uses it, so differences between methods reflect how they decide, not which model they use. 
We use a compact model because each result is repeated over 20 seeds, three \support levels and three training losses, with collection and retraining on top; we check that the findings hold at other sizes (7M and 115M) and on other backbones as detailed in the supplement.
We compare against world-action models and policy learning on world models (\cref{sec:strategy_results}).

\paragraph{Evaluation.} Decisions use the planner $\pi_\beta$ of \cref{sec:planners}, and regret is normalized so that a random action scores 1 and the best action in the bank 0. Each step of \CURE is evaluated with the training data, test set, seeds and planner held fixed.
For Certify, we test whether the certificate of each decision predicts its regret, and compare it with video quality scores.
For Upweight, we compare the regret of each world model across \support levels.
For Rollout, we compare regret as the number of candidate actions scored grows, at each \support level.
For Explore, we give each collection rule the same budget of new clips, retrain the model, and compare the resulting regret.
Implementation details, per-domain choices, and results are in the supplementary material.

\subsection{Certify (D1)}\label{sec:trust_results}

\begin{figure}[!tbp]
  \centering
  \includegraphics[width=\linewidth]{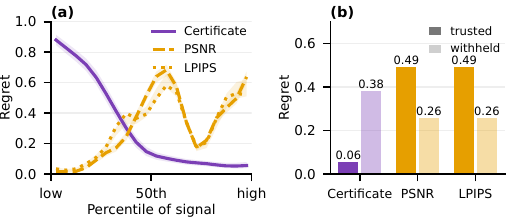}
  \caption{\textbf{The certificate predicts decision error while video quality does not.} Setting: Language-Table sim. (a)~Regret against the percentile of each signal (smoothed; 95\% bands). LPIPS is negated, so higher means better video. (b)~Regret of trusted and withheld decisions. PSNR and LPIPS trust the same share as the certificate.}
  \label{fig:trust}
\end{figure}

We test whether the certificate from \cref{eq:certificate} separates good decisions from bad ones, and whether video quality does too.
In \cref{fig:trust}a, we sort decisions by each signal (certificate $\dxScore$, PSNR, LPIPS), from least to most trusted (low to high $\dxScore$ and PSNR, high to low LPIPS). We then plot the mean regret of decisions at each percentile of each signal.
We find that regret falls steadily as the certificate rises (Spearman $\rho=-0.67$, 95\% interval $[-0.69,-0.65]$), meaning a higher certificate reliably identifies a better decision. 
Video quality does not: among the decisions of any one trained model, PSNR is unrelated to regret ($\rho\approx-0.03$), while the certificate's correlation is $-0.40$ (negative for every model).

In \cref{fig:trust}b, we use each signal to plan. 
Decisions that pass the certificate are almost always good: they make up 22\% of decisions, with regret $0.06$, compared to $0.38$ for the decisions it does not certify. If these decisions were made based on the equivalent top 22\% of videos according to PSNR, those decisions are worse than the rest, with regret $0.49$ against $0.26$.

Across domains\footnote{Please refer to the supplementary material for per-domain results}, the certificate finds where the training data lacks \support and ranks world models by regret, which video quality does not. The certificate also separates good decisions from bad: those it certifies have lower regret than those it withholds. 
Across whole models, Fréchet Video Distance (FVD)~\citep{unterthiner2018fvd} tells a similar story: on held-out clips of logged actions, averaged over \support levels, the standard model's videos are more realistic than \CURE's (FVD $22$ vs.\ $80$), but regret is $3.1\times$ higher.

\subsection{Explore (D2)}\label{sec:collect_results}

% \begin{figure}[!tbp]
%   \centering
%   \includegraphics[width=\linewidth]{F5_collection_a}\\[2pt]
%   \includegraphics[width=\linewidth]{F5_collection_b}
%   \caption{\textbf{How and where to collect.}
%   (a) Regret after 300 new clips at low \support, by strategy; numbers are paired differences from \CURE.
%   (b) Regret averaged over \support levels against total clips; rings mark the oracle allocation.}
%   \label{fig:collection}
% \end{figure}

\emph{How to collect.} We compare collection strategies at low \support with a fixed budget of 300 new clips. \CURE collects where \support is missing and more than halves regret ($0.73$ to $0.29$).
Here clips are allocated across \support levels, but any grouping of scenes, such as by instruction or wind condition, can be used.
Rules that follow the behavior policy do worse: more clips from the same behavior policy give $0.38$, and clips sampled from the planner $\pi_\beta$ with $\beta>0$, which stays close to the behavior policy, give $0.41$, against $0.30$ for \CURE's rule in this comparison.
Rules that follow the model also do worse: collecting the action it ranks best in each scene gives $0.34$, and collecting where it is most uncertain gives $0.38$.
Collecting uniformly over actions does well here ($0.31$), but only because at low \support almost every action needs data. Uniform collection does not know where \support is missing, which matters once the budget must be split across \support levels as shown below.

\emph{Sample efficiency.} \CURE spends its clip budget where the certificate is weakest first, in increasing order of the mean certificate across \support levels. We compare it with splitting the budget evenly across \support levels (even split), at equal total clips. 
The even split is less efficient than \CURE: with 300 clips its regret is higher by $0.020$ (paired 95\% interval $[0.008, 0.032]$), and it needs 900 clips to reach what \CURE reaches with a budget of 600 ($0.13$).

The same story holds across domains: collecting where \support is missing lowers regret at low \support in every domain except MuJoCo smooth, the control with nothing to fix, and spreading the same clips evenly across \support levels is worse than \CURE.

\subsection{Upweight (D3)}\label{sec:upweight}

We compare the standard, debiased, and debiased with retargeting (just called retargeted) models across \support levels on Language-Table sim.
At high \support, every model is already near zero regret --- weights are stable, and there are enough demonstrations to plan well.
At partial \support, debiasing lowers regret by about a third (from $0.25$ to $0.17$), and retargeting lowers it further ($0.14$).
At low \support, every model fails equally ($0.72$ to $0.73$), since no loss can recover actions the training data never shows; this is where new data is needed (\cref{sec:collect_results}).
Removing the Upweight step from \CURE confirms this (\cref{tab:strategy}): averaged over domains, regret at partial \support rises from $0.18$ to $0.24$, and from $0.11$ to $0.12$ at high \support.

Two findings from the other domains qualify this picture. 
Low \support is not always hopeless: at equal training data, upweighting lowers regret at low \support by $0.23$ on \kickc and by $0.13$ on Language-Table real.
Also, debiasing is not always the right call. In particular, when the weights add variance with little reduction in bias, \CURE keeps the standard model, using a check on the validation set.
For example, on MuJoCo contact debiasing raises regret at partial and high \support (Supplement), and \CURE selects the standard model.

\subsection{Rollout (D4)}\label{sec:compute_results}

\begin{figure}[!tbp]
  \centering
  \includegraphics[width=\linewidth]{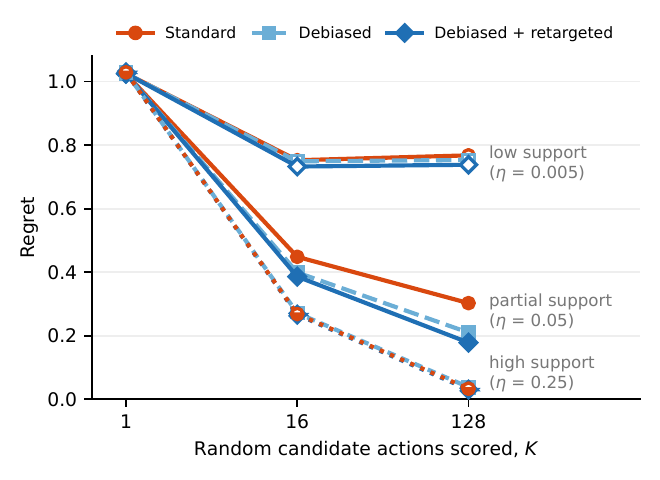}
  \caption{\textbf{Test-time compute helps only where the model has \support.} Regret against number of candidates scored $K$. Hollow: low; filled: partial; dotted: high \support.}
  \label{fig:search}
\end{figure}

Test-time compute helps only where the model has \support, and the certificate tells \CURE where that is (\cref{fig:search}). At high \support, all methods are competitive. At partial \support, scoring 128 candidate actions instead of 16 halves the regret of the retargeted model ($0.39$ to $0.18$) and helps the standard model less ($0.45$ to $0.30$). At low \support, regret stops improving after 16 candidates ($0.75$ to $0.77$ for the standard model), since the issue is not variance, it is bias. Averaging more rollouts per action behaves the same way: going from 1 to 16 lowers the retargeted model's regret at partial \support from $0.17$ to $0.12$, but leaves low-\support regret above $0.68$. The certificate separates these cases: it passes $4\%$ of decisions at low \support, against $25\%$ at partial and $37\%$ at high \support. \CURE therefore spends compute where the certificate finds \support, and collects new clips where it does not.

\subsection{\CURE as a strategy}\label{sec:strategy_results}

Across five domains, the certificate predicts decision error where video quality does not; collecting where \support is missing is the main fix at low \support, debiasing at partial \support, and extra test-time compute helps only once \support exists. Choosing among these fixes by the certificate lowers regret over standard world models by 31--61\% at low \support and up to 50\% at partial \support, without raising it at high \support. Against baselines from every family, given the same data and compute, \CURE has the lowest or tied-lowest regret in every domain.
We evaluate \CURE holistically, first across differing \support levels, and in comparison to competing baselines mentioned in \cref{sec:planners}.
Recall that low, partial and high \support mean that a fraction $\eta$ of $0.5\%$, $5\%$ and $25\%$ of the behavior policy's clips depart from its default action. 
\begin{table}[!tbp]
  \centering\small
  \setlength{\tabcolsep}{4pt}
\caption{\textbf{\CURE has the lowest regret at every \support level.} Regret (lower is better), averaged over \kickc, Language-Table sim and real, and MuJoCo contact, with 95\% intervals. All rows share the model, data (incl.\ \CURE's 300 clips at low \support) and evaluation. Rollout uses 16 rollouts per action; $^\dagger$4, as for the baselines. Gray: step not called, so equal to \CURE. $^\ddagger$Uses ten world models instead of one. Baselines are described in \cref{sec:planners}; per-domain results are in the supplement.}
\resizebox{\linewidth}{!}{%
  \begin{tabular}{@{}lccc@{}}
    \toprule
    & low & partial & high \\
    \midrule
    \multicolumn{4}{@{}l}{\emph{One fix alone (no certificate)}} \\
    \quad Standard (no fix) & 0.693\,{\scriptsize$\pm$0.024} & 0.257\,{\scriptsize$\pm$0.016} & 0.126\,{\scriptsize$\pm$0.007} \\
    \quad Upweight only & 0.626\,{\scriptsize$\pm$0.040} & 0.253\,{\scriptsize$\pm$0.063} & 0.173\,{\scriptsize$\pm$0.050} \\
    \quad Rollout only & 0.646\,{\scriptsize$\pm$0.024} & 0.240\,{\scriptsize$\pm$0.016} & 0.115\,{\scriptsize$\pm$0.006} \\
    \quad Explore only & 0.436\,{\scriptsize$\pm$0.019} & 0.257\,{\scriptsize$\pm$0.016} & 0.126\,{\scriptsize$\pm$0.007} \\
    \midrule
    \multicolumn{4}{@{}l}{\emph{Ablation: \CURE without one step}} \\
    \quad without Explore & 0.605\,{\scriptsize$\pm$0.019} & \textcolor{gray}{0.178\,{\scriptsize$\pm$0.008}} & \textcolor{gray}{0.107\,{\scriptsize$\pm$0.005}} \\
    \quad without Upweight & 0.436\,{\scriptsize$\pm$0.019} & 0.240\,{\scriptsize$\pm$0.016} & 0.115\,{\scriptsize$\pm$0.006} \\
    \quad without Rollout$^\dagger$ & \textcolor{gray}{0.385\,{\scriptsize$\pm$0.021}} & 0.192\,{\scriptsize$\pm$0.010} & 0.121\,{\scriptsize$\pm$0.004} \\
    \midrule
    \multicolumn{4}{@{}l}{\emph{Baselines}$^\dagger$} \\
    \quad Behavior policy & 0.991\,{\scriptsize$\pm$0.004} & 0.959\,{\scriptsize$\pm$0.009} & 0.985\,{\scriptsize$\pm$0.005} \\
    \quad World-action model & 0.845\,{\scriptsize$\pm$0.035} & 0.790\,{\scriptsize$\pm$0.025} & 0.458\,{\scriptsize$\pm$0.010} \\
    \quad Search with a verifier & 0.705\,{\scriptsize$\pm$0.010} & 0.520\,{\scriptsize$\pm$0.012} & 0.250\,{\scriptsize$\pm$0.006} \\
    \quad TD3+BC & 0.657\,{\scriptsize$\pm$0.013} & 0.497\,{\scriptsize$\pm$0.014} & 0.337\,{\scriptsize$\pm$0.005} \\
    \quad AWR & 0.751\,{\scriptsize$\pm$0.010} & 0.710\,{\scriptsize$\pm$0.010} & 0.410\,{\scriptsize$\pm$0.009} \\
    \quad BCQ & 0.909\,{\scriptsize$\pm$0.005} & 0.834\,{\scriptsize$\pm$0.009} & 0.677\,{\scriptsize$\pm$0.006} \\
    \quad SPIBB & 0.656\,{\scriptsize$\pm$0.009} & 0.573\,{\scriptsize$\pm$0.010} & 0.487\,{\scriptsize$\pm$0.006} \\
    \quad Count lower bound & 0.495\,{\scriptsize$\pm$0.017} & 0.297\,{\scriptsize$\pm$0.012} & 0.152\,{\scriptsize$\pm$0.005} \\
    \quad Mean $-\,\lambda\cdot$sd & 0.462\,{\scriptsize$\pm$0.018} & 0.276\,{\scriptsize$\pm$0.015} & 0.146\,{\scriptsize$\pm$0.005} \\
    \quad Worst rollout & 0.460\,{\scriptsize$\pm$0.017} & 0.279\,{\scriptsize$\pm$0.015} & 0.146\,{\scriptsize$\pm$0.005} \\
    \quad Ensemble lower bound$^\ddagger$ & 0.428\,{\scriptsize$\pm$0.011} & 0.236\,{\scriptsize$\pm$0.003} & 0.127\,{\scriptsize$\pm$0.003} \\
    \midrule
    \textbf{\CURE} & \textbf{0.385}\,{\scriptsize$\pm$0.021} & \textbf{0.178}\,{\scriptsize$\pm$0.008} & \textbf{0.107}\,{\scriptsize$\pm$0.005} \\
    \bottomrule
  \end{tabular}}
\label{tab:strategy}
\end{table}

\emph{\CURE across \support levels.} \Cref{tab:strategy} compares \CURE with each fix applied alone at every \support level, with \CURE missing one step, and with baselines from each family.
At low \support, only new clips help; the certificate fails, and \CURE collects.
At partial \support, upweighting and averaging more rollouts help; the certificate passes, and \CURE acts with both.
At high \support, the model is already close to correct, and extra rollouts trim what remains.
\CURE has the lowest regret at every \support level: $0.39$, $0.18$ and $0.11$, against $0.69$, $0.26$ and $0.13$ for the standard model and at least $0.44$, $0.24$ and $0.12$ for any single fix.
Removing one step at a time shows where each matters: without Explore, regret at low \support rises to $0.60$; without Upweight, regret at partial \support rises to $0.24$; and without Rollout, regret at partial and high \support rises to $0.19$ and $0.12$. Where the certificate does not call for a step, removing it changes nothing. 
Upweight and Rollout help more together, since extra rollouts reduce variance but not bias.

\emph{\CURE against baselines.} Given the same data, \CURE's regret is below that of world-action models, search with a verifier and policy learning in the world model at every \support level.
The strongest baselines are decision rules that score actions pessimistically, by a lower bound or the worst imagined rollout.
\CURE's regret is lower than all of them at every \support level, by 16--36\% for the single-model rules. An ensemble lower bound, which runs ten world models, comes closest in regret ($0.428$, $0.236$ and $0.127$ against \CURE's $0.385$, $0.178$ and $0.107$). They act on the decision rule rather than the model or data, so they are complementary to \CURE.

\CURE's regret gains, largest at low \support, hold beyond our 28M model: at 115M, 340M, on DIAMOND~\cite{alonso2024diamond}, and on the pretrained CogVideoX-2B~\cite{cogvideox2024}. On CogVideoX-2B the certificate still tracks decision quality: its rank correlation with regret is $-0.38$ within each \support level and seed, and certified decisions have regret $0.016$ against $0.245$ for the rest (Supplement).
% Collecting 300 clips at every \support level is competitive with our collection strategy. 
% But it spends three times \CURE's budget, including where the certificate already passes. Spread over the same 300 clips, collecting at every level is worse than \CURE in every domain where we ran it, and given a larger budget, \CURE's rule reaches the same regret with fewer clips (\cref{sec:collect_results}).
\section{Conclusions}\label{sec:limits}
World model visual quality says little about how well it can plan. While several solutions exist to close this gap, there is little guidance on which solution to use. We introduce \CURE, which certifies when a world model can be trusted for planning. When it cannot, \CURE recommends when to upweight the training data, when to roll out more at test time, and when and where to explore.
As world models are increasingly used for planning, \CURE provides a certificate-guided framework for assessing when an existing model can be trusted and deciding how to allocate limited data and compute where they matter.
%As world models are increasingly used to plan, \CURE tells practitioners when an existing model can be trusted and offers a certificate-guided framework for spending limited data and compute where they matter.

% We acknowledge limitations of our study. First, we study single decisions, and leave multi-step decision problems and unobserved confounding for future work. Second, in our experiments counterfactual outcomes are available only in simulation (\kickc, MuJoCo, Language-Table sim). On real video, untaken actions are scored by a referee model. Finally, the certificate is conservative, and future work can sharpen the bound.

% limited regimes, we create a technique that does Blah for general soln. 

\clearpage 

\section*{LLM Usage Statement}
We used Claude (Anthropic) as an assistant throughout the preparation of this paper.
\textbf{Writing:} it drafted, edited and shortened text in all sections of the main paper and the supplement, from the authors' outlines, notes and results; the authors revised and approved all text.
\textbf{Figures:} it wrote plotting code for the schematic figures Fig.~\ref{fig:tree} and for the layout and styling of the result figures; all plotted values come from the authors' experiment outputs.
\textbf{Analysis and checking:} it helped draft experiment specifications and analysis plans, checked the consistency of reported numbers against result tables, and reviewed the paper for errors.
The authors take full responsibility for the content of the paper.

{
    \small
    \bibliographystyle{ieeenat_fullname}
    \bibliography{main}
}

% Supplement: submit separately per CVPR rules; kept inline here for drafting.
% \clearpage
% \input{sec/X_suppl}
\end{document}